# Two-Stage Fine-Tuning: A Novel Strategy for Learning Class-Imbalanced Data

Taha ValizadehAslani[a], Yiwen Shi[b], Jing Wang[c], Ping Ren[c], Yi Zhang[c], Meng Hu[c], Liang Zhao[c], Hualou Liang[d*]

[a] Department of Electrical and Computer Engineering, College of Engineering, Drexel University, Philadelphia, PA, USA
[b] College of Computing and Informatics, Drexel University, Philadelphia, PA, USA
[c] Office of Research and Standards, Office of Generic Drugs, Center for Drug Evaluation and Research, United States Food and Drug Administration, Silver Spring, MD, USA
[d] School of Biomedical Engineering, Science and Health Systems, Drexel University, Philadelphia, PA, USA

*Corresponding author: hualou.liang@drexel.edu

**Abstract:** Classification on long-tailed distributed data is a challenging problem, which suffers from serious class-imbalance and hence poor performance on tail classes with only a few samples. Owing to this paucity of samples, learning on the tail classes is especially challenging for the fine-tuning when transferring a pretrained model to a downstream task. In this work, we present a simple modification of standard fine-tuning to cope with these challenges. Specifically, we propose a two-stage fine-tuning: we first fine-tune the final layer of the pretrained model with class-balanced reweighting loss, and then we perform the standard fine-tuning. Our modification has several benefits: (1) it leverages pretrained representations by only fine-tuning a small portion of the model parameters while keeping the rest untouched; (2) it allows the model to learn an initial representation of the specific task; and importantly (3) it protects the learning of tail classes from being at a disadvantage during the model updating. We conduct extensive experiments on synthetic datasets of both two-class and multi-class tasks of text classification as well as a real-world application to ADME (i.e., absorption, distribution, metabolism, and excretion) semantic labeling. The experimental results show that the proposed two-stage fine-tuning outperforms both fine-tuning with conventional loss and fine-tuning with a reweighting loss on the above datasets.

## 1. Introduction

Real-world classification problems typically exhibit an imbalanced or long-tailed distribution (Deng et al., 2009; Zhu et al., 2014; Horn and Perona, 2017; Buda et al., 2017; Cui et al., 2018; Johnson and Khoshgoftaar, 2019; Liu et al., 2019; Madabushi et al., 2020; Jamal et al., 2020; Yang and Xu, 2020), wherein some classes have fewer samples (minority classes) and some classes have more samples (majority classes). Such imbalance poses a significant challenge for many machine learning algorithms (Collobert and Weston, 2008; He and Garcia, 2009; Horn and Perona, 2017; Buda et al., 2017; Ando and Huang, 2017; Yang and Xu, 2020; Madabushi et al., 2020). Poor performance of the model on the minority classes during the training can have a negative impact on the overall performance when these classes are represented in the test set, or if the chosen performance metric places equal importance on all classes, regardless of their class frequency. The latter can be done because of various practical concerns, such as fairness or transferability to new domains (Cao et al., 2019). Moreover, the minority classes might have intrinsic importance, despite being rare (He and Garcia, 2009; Branco et al., 2016). Hence, a solution that improves the performance of the minority classes, while maintaining the good performance of the majority classes is of crucial importance.

Fine-tuning is the prevalent paradigm for using large pretrained language models (LMs) (Radford et al., 2019; Devlin et al., 2019; Raffel et al., 2020) to perform downstream tasks. This paradigm, while widely used, is far from ideal, as the fine-tuning may predispose the pretrained model to overfitting and generalizing poorly to out-of-distribution (OOD) data due to the huge model and relatively small size of the domain-specific data (Zhao et al., 2019; Guo et al., 2020; Radiya-Dixit and Wang, 2020; Gordon et al., 2020; Zaken et al., 2021). This is consistent with the theoretical analysis that fine-tuning can distort the pretrained features and lead to poor OOD accuracy (Kumar et al., 2022). In a class-imbalance scenario, the model tends to over-fit the minority classes due to their limited representation, even with the specialized class-balanced techniques such as re-sampling (Buda et al., 2017; He & Garcia, 2009; Van Horn & Perona, 2017) and re-weighting (Cao et al., 2019; Dong et al., 2017; Huang et al., 2016; Cui et al., 2019a).

Empirically, it is showed that the reweighting has a significant effect *early in training*, and the impact of importance weighting diminishes over successive epochs of training (Byrd & Lipton, 2019). Theoretical analysis of imbalanced learning (Fang et al., 2021) predicts that the minority classes collapse to a single vector *in the topmost layer*, which places a fundamental limit on the model performance for the minority classes. Motivated by the empirical findings and theoretical results, we propose a two-stage fine-tuning to first fine-tune the final layer of the pretrained model with reweighting to allow the model to learn an initial representation that is fair for each class, and then perform the standard or vanilla fine-tuning. Our proposed method is schematically shown in Figure 1(b), as compared to the vanilla fine-tuning shown in Figure 1(a).

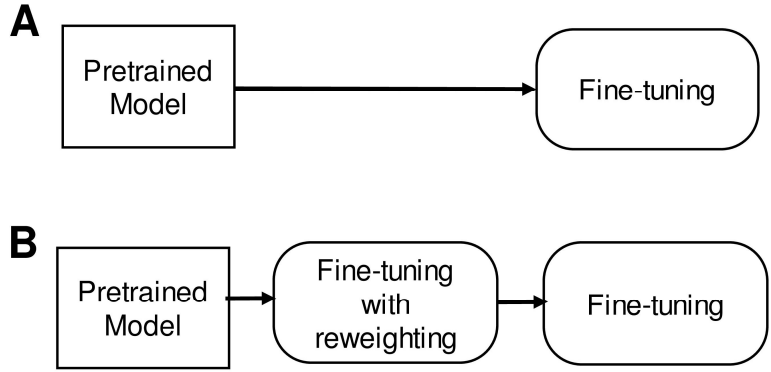

**Fig.1 a** Vanilla fine-tuning: As a predominant way to conduct model adaptations, fine-tuning initializes with the general-purpose model pretrained with a large unlabeled corpus, and then performs a small amount of task-specific parameter updates for the task of interest. **b** The two-stage fine-tuning we proposed to improve imbalanced classification: (1) fine-tuning with reweighting, and (2) standard fine-tuning. The first stage allows the model to learn a good initial representation for the second stage and protect the learning of minority classes from being at a disadvantage during the model updating.

The rest of this paper is organized as follows. In Section 2, we briefly review the related work. Then, in Section 3, we delineate the details of our proposed two-stage fine-tuning. In Section 4, we describe the experiments and demonstrate the results on both synthetic datasets and real-world application, which show that our two-stage fine-tuning outperforms the vanilla fine-tuning and the fine-tuning only with a reweighting loss for both in-distribution (ID) and OOD test data. Section 5 provides the discussions and the conclusion.

## 2. Related Work

We first provide preliminaries of the prevalent pretraining then fine-tuning paradigm in modern Natural Language Processing (NLP). We then introduce resampling, reweighting and ensemble-based approaches that are widely used to alleviate the class-imbalanced problem in long-tailed datasets (e.g., Japkowicz and Stephen 2002; Bengio, 2015; Geifman & El-Yaniv, 2017; Huang et al., 2016; Van Horn & Perona, 2017; Cui et al., 2019; Krawczyk 2016; He and Garcia, 2009).

**Pretraining and Fine-tuning Paradigm**. Large language models (LMs) have been shown to achieve remarkable performance across a variety of natural language tasks (Devlin et al., 2019; Radford et al., 2019; Raffel et al., 2020), which ushers in a new era of pretraining and fine-tuning. In this paradigm, a large LM such as BERT, which stands for Bidirectional Encoder Representations from Transformers, is trained on vast amounts of text, then fine-tuned on a specific downstream task. Among different fine-tuning approaches, vanilla fine-tuning is perhaps the most popular approach, which fine-tunes some or all the layers of the LM and then adds one or two simple task-specific output layers (known as prediction heads, Wolf et al., 2020). In this work, we present a simple modification of vanilla fine-tuning to cope with imbalanced data.

**Resampling.** It includes either over-sampling the minority classes or under-sampling the majority classes, or both. Such re-sampling incurs the cost of overfitting or losing the important information respectively. In addition, new samples can also be generated by interpolating neighboring samples or synthesizing for minority classes (Chawla et al., 2003; He & Garcia, 2009).

**Reweighting**. This approach involves adjusting the loss function to compensate for class imbalance by assigning weights to different samples according to the class distribution (Cao et al., 2019; Cui et al., 2019; Dong et al., 2017; Huang et al., 2016). There are different importance weighting schemes. A simple way is to assign sample weights proportionally to the inverse of the class frequency (Huang et al., 2016; Wang et al., 2017). Such a scheme tends to perform poorly when training on large-scale, imbalanced datasets (Huang et al., 2016). Rather than using the total number of samples present in each class, re-weighting loss by the inverse effective number of samples is introduced for better class-balance (Cui et al., 2019). In addition to adjusting weights on class level, re-weighting can also be performed on sample level by increasing the weights for hard samples and decreasing the weights for easy samples (Dong et al. 2017; Lin et al. 2017; Li et al. 2019). The downside for assigning higher weights to hard examples is the overstress on noisy or mislabeled samples (Koh and Liang 2017; Ren et al. 2018). Recently, margin approaches such as the label-distribution-aware margin (LDAM) have been proposed to encourage the minority classes to have larger margins, leading to competitive performance on many benchmark vision tasks (Cao et al., 2019). We benchmark our two-stage fine-tuning against this method.

**Ensemble-based Approaches**. Ensemble imbalanced learning is known to effectively deal with imbalanced data by merging the outputs of multiple classifiers (Liu et al., 2020; Chawla et al., 2003; Wang and Yao, 2009) or by combining individual classifiers in a multi-expert framework (Wang et al. 2021). These methods achieve the state-of-the-art performance mainly by reducing the model variance to obtain robust predictions (Krawczyk, 2016). However, many of them are direct combinations of a resampling/reweighting scheme and an ensemble learning framework (e.g. Chawla et al., 2003), which hence inherit the similar shortcomings of existing class-balancing strategies, such as the redundancy for easy samples by uniformly assigning experts to all classes (Wang et al. 2021). In most cases, it is difficult to obtain an appropriate cost matrix given by domain experts (Krawczyk et al., 2014).

# 3. Methods

## 3.1 Motivations

Our work draw inspiration primarily from recent theoretical work and empirical findings when training deep learning models on imbalanced datasets. While empirically the class imbalance often leads to inferior model performance (Buda et al., 2017; He and Garcia, 2009, Van Horn and Perona, 2017), there remains a lack of theoretical understanding of its impact on the deep models. Recent theoretical model (Fang et al. 2021) predicts the emergence of Minority Collapse in imbalanced learning, i.e., the minority classes collapse to a single vector in the topmost layer, which places a fundamental limit on the model performance for the minority classes. Further mathematical analysis reveals that fine-tuning of the entire model can distort the pretrained features and underperform on test examples sampled from OOD distributions (Kumar et al., 2022). These theoretical studies signify that only the last few layers actually need to be fine-tuned when it comes to imbalanced datasets.

The key empirical motivation for our method rested upon recent experimental work (Byrd and Lipton 2019) for studying the effects of importance weighting in deep learning. Across a variety of tasks, architectures, and data sets, they found that the weighting has a significant effect *early in training*, and the impact of importance weighting diminishes over successive epochs of training. These results provide the compelling evidence that *the early stage of training* is important to imbalanced learning. Therefore, we are motivated to add a new learning stage where the final layer of the model is trained before the standard fine-tuning takes place.

## 3.2 Two-stage Fine-tuning

In order to maintain the good overall performance for both the majority classes and the minority classes, we propose a two-stage fine-tuning approach to improving imbalanced classification:

**Stage 1: Fine-tuning with reweighting.** Given a pretrained model (e.g., BERT), we first train the model with a class-balanced reweighting loss function, which assigns larger weights to the minority classes, so that the model fits the minority classes, in spite of their limited samples. In order to avoid distorting the pretrained features, we only train the final layer of the model and the classification head. The first stage allows the model to learn a good initial representation of the specific downstream task and protects the learning of minority classes from being at a disadvantage. While this algorithm can work with any reweighted loss function, in our implementation, we use the LDAM (Cao et al., 2019) loss, which encourages the model to gain the optimal trade-off between per-class margins.

**Stage 2: Standard fine-tuning.** In this stage, we simply fine-tune the entire model, where the conventional cross-entropy loss function is used.

The pseudo-code for our proposed method is provided in Algorithm 1.

---
**Algorithm 1** Two-stage Fine-tuning
---
**Require**: A pretrained model
 1: Load the model
 2: *# Stage 1: Fine-tuning with reweighting*
 3: Freeze every layer except for the final layer and the classification head
 4: **for** *i* = 1*...NumEpoch*1 **do**
 5:     Fine-tune the model on all samples in the train set with class-balanced reweighing loss function
 6: **end for**
 7: Unfreeze the frozen layers
 8: *# Stage 2: Standard fine-tuning*
 9: **for** *i* = 1*...NumEpoch*2 **do**
10:     Fine-tune the model on all samples in the train set with conventional loss function
11: **end for**
12: *# Test*
13: Test the model on the test set
---

## 4. Experiments

### 4.1 General Settings

We conduct our experiments on two synthetic benchmarks and one real-world application. The benchmark datasets are The Stanford Sentiment Treebank-2 (SST-2) (Socher et al., 2013) which is a two-class sentiment classification task, and the 20 Newsgroups (Lang, 1995) which is a multi-class text classification. We artificially create the different imbalanced versions of two benchmark datasets by following specific imbalance distributions, as detailed below. Our real-world application is the semantic classification of pharmacokinetics paragraphs in drug labels into five classes, namely, absorption, distribution, metabolism, excretion (ADME), and other (Shi et al., 2022b). This dataset is naturally imbalanced.

In all our implementations, we use the BERT model (*bert-base-uncased*[1]) as the pretrained model which was initialized with the parameters released by (Devlin et al., 2019), and can be accessed from Huggingface (Wolf et al., 2020). All the results we reported are the average of the five runs, each with a different seed. All the experiments were run on Nvidia Tesla V100-SXM2-32GB.

#### 4.1.1 Baselines

For each dataset, we benchmark our proposed method against two baseline methods (vanilla fine-tuning and LDAM fine-tuning): (1) Vanilla fine-tuning: This is just the standard fine-tuning of the entire model, where a conventional cross-entropy loss function is used. Here, no measure is taken against the imbalanced data; and (2) Fine-tuning with LDAM loss: This method is the same as the vanilla fine-tuning, except we use the LDAM loss function due to its excellent performance (Cao et al., 2019).

---
[1] https://huggingface.co/bert-base-uncased

### 4.1.2 Hyperparameter tuning

We tune the hyperparameters for the vanilla method by sweeping over all the combinations of epoch number, $e \in \{1, 2, ..., 6\}$, and learning rate, $LR \in \{10^{-5}, 5 \times 10^{-5}, 10^{-4}, 5 \times 10^{-4}\}$. For ADME and SST-2, we set the maximum sequence length to 128 because of the short text length in each sample, and the batch size to 16. For the 20 Newsgroups, we set the maximum sequence length to 512 due to its relatively long text length, and the batch size to 8 to avoid running into memory issues.

For LDAM fine-tuning, we use the same hyperparameters as the vanilla fine-tuning. For our two-stage fine-tuning, in Stage 1, we use the learning rate of $10^{-4}$, set the number of epochs to 3 for ADME and Newsgroup, and 1 for SST-2. A smaller epoch number is used for SST-2 because this dataset has a large number of training samples compared to the other datasets. For Stage 2, we use the same hyperparameters as the vanilla fine-tuning.

### 4.1.3 Evaluation metrics

To evaluate the performance of the model, we use the F1-score that is defined as the harmonic mean of the precision and recall. For both two-class and multi-class problems, we report both the overall F-1 score (micro-F1) and per-class F-1 score to quantify the generalization performance of both majority and minority classes for the benchmark datasets. The metrics were calculated using scikit-learn library (Pedregosa et al., 2011).

## 4.2 Synthetic Benchmark 1: SST-2

### 4.2.1 Dataset

The Stanford Sentiment Treebank (Socher et al., 2013) is a corpus of sentences from movie reviews and human annotations of their sentiments, which now is one of the General Language Understanding Evaluation (GLUE) benchmark (Wang et al., 2018). In this dataset, each review is labeled either 1 for positive sentiment, or 0 for negative sentiment. There are 37569 positive reviews and 29780 negative reviews for the train data, and almost evenly distributed test data (444 positive and 428 negative).

To create the imbalanced training set, we manually remove a number of negative reviews until a given imbalance ratio is reached. Here we define the imbalance ratio as the ratio of the number of the minority samples to that of the majority samples. For instance, the imbalanced ratio of 0.2 denotes that the number of the negative samples is one-fifth of the positive samples. The testing dataset remained unchanged.

### 4.2.2 Results

We first show our main finding with the imbalanced ratio of 0.2. Results for different methods are presented in Table 1. The best performance is achieved by our two-stage fine-tuning, where the F1 scores are improved for both classes, but importantly more improvement is observed for the minority class than the majority class.

**Table 1** Comparison of per-class and micro F1 between different methods on the SST-2 dataset when the imbalanced ratio is set to 0.2. The values in parenthesis show the F1 improvements over vanilla fine-tuning. The best performance is achieved by our two-stage fine-tuning.

|  | Micro F1 (Improvement) | F1 (Improvement) negative class | F1 (Improvement) positive class |
| --- | --- | --- | --- |
| **Vanilla** | 0.8823 | 0.8698 | 0.8926 |
| **LDAM** | 0.8865 (0.0042) | 0.8750 (0.0052) | 0.8960 (0.0034) |
| **Two-stage** | **0.8956 (0.0133)** | **0.8859 (0.0161)** | **0.9039 (0.0113)** |

**Effect of imbalance ratio**. To examine the effect of imbalance ratio on the model performance, we systematically vary the imbalance ratio from 0.1 to 0.7, with increments of 0.1. Note we can only go up to 0.7 due to the relative imbalance in the original data.

Figure 2 shows the comparison of three methods (Vanilla, LDAM and our two-stage fine-tuning) in terms of how the F1 score changes as a function of the imbalance ratio, where the micro F1s are shown in Figure 2(a) and the per-class F1 in Figure 2(b). Among the three methods, our two-stage fine-tuning performs the best, followed by LDAM, and the vanilla method has the worst performance in F1 score. This is especially evident at the extremely imbalanced ratio of 0.1 (the leftmost points on both panels) where the F1 score of the vanilla method is rather poor for the minority class. This result is not unexpected as the vanilla method does not compensate for the imbalanced data. Critically, when comparing three methods, we observe from Figure 2(b) that in our two-stage fine-tuning, there is a larger F1 improvement for the minority class (negative) relative to the majority class (positive). Such improvement diminishes as the imbalance ratio increases. At the imbalance ratio of 0.7 (the rightmost point in both panels) where the imbalance becomes less of a problem, all the methods have reached the similar performance.

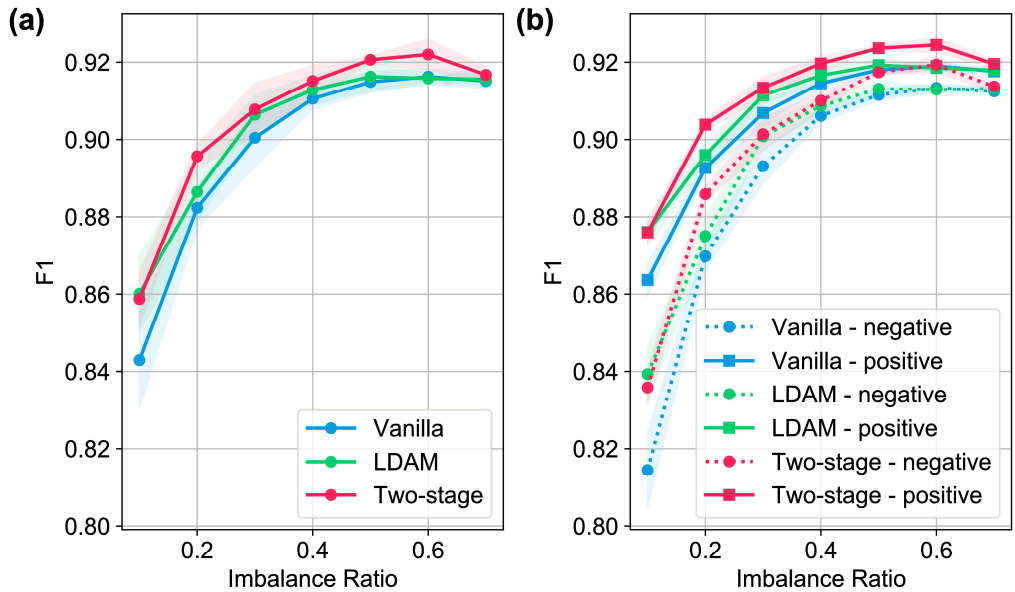

**Fig.2** Impact of imbalance ratio on the model performance of F1 score in the SST-2 dataset across three methods: vanilla, LDAM and two-stage fine-tuning. **a** Micro F1 changes as a function of imbalance ratio. **b** Per-class F1 changes as a function of imbalance ratio. It is observed that our two-stage fine-tuning outperforms both the vanilla and LDAM methods in both micro F1 and per-class F1. Importantly, among the three methods, there is a large F1 improvement for the minority class (negative) relative to the majority class (positive). Such improvement diminishes as the imbalance ratio increases. Shaded areas represent standard error.

**Out-of-distribution samples**. The OOD samples are the data unseen during training, which usually have a different distribution than the training data. This is known as the distribution shift (Zhang et al., 2021), which is a fundamental challenge in machine learning. As such, the OOD data provides a critical test for the generalization ability of new method. We therefore check this idea using the models trained by the SST-2 data and tested on a different sentiment analysis dataset: Internet Movie Database (IMDB) reviews (Maas et al., 2011). The IMDB dataset contains 25,000 test samples with an equal number of positive and negative reviews. Figure 3 shows the testing results of IMDB data. From Figure 3, we observe overall that our two-stage fine-tuning performs better than the vanilla method on OOD test data. This is expected as the simple modification we made to standard fine-tuning at Stage 1 is to only fine-tune a small portion of the model parameters while keeping the rest untouched; it hence can preserve pretrained features, which allow our method to extrapolate better to OOD data. We note our OOD results are also compatible with the theoretical results by showing that fine-tuning the full model can distort the pretrained features and lead to poor OOD accuracy (Kumar et al., 2022).

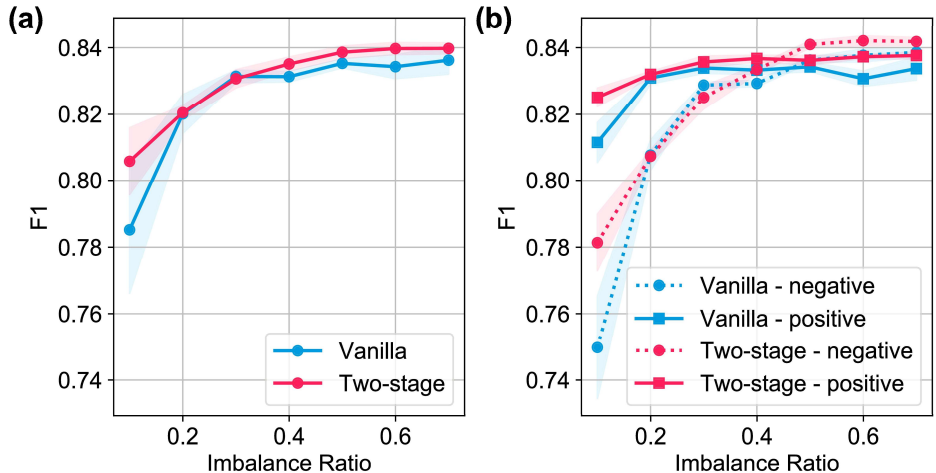

**Fig.3** The F1 scores vary with the imbalance ratio for OOD testing for both vanilla method and our two-stage fine-tuning. The models are trained on SST-2 and tested on IMDB at different imbalance ratios. **a** Micro F1 changes as a function of imbalance ratio. **b** Per-class F1 changes as a function of imbalance ratio. Overall, our two-stage fine-tuning has a better generalization ability than the vanilla method for most of the OOD testing. Shaded areas represent standard error.

### 4.3 Synthetic Benchmark 2: The 20 Newsgroups

#### 4.3.1 Dataset

The 20 Newsgroups dataset (Lang, 1995) is a collection of approximately 20,000 newsgroup documents, partitioned almost evenly across 20 different newsgroups. We manually create the imbalanced version of the training set by reducing the number of training examples per class and keep the test set unchanged. Two types of imbalance are considered to ensure that our method applicable to different settings. One is the step imbalance (Buda et al., 2017), where we select ten classes to reduce their sample sizes to 59, which are approximately one-tenth of the sample sizes of the remaining classes that are between 591 and 600. As a result, all minority classes have the same sample size, as roughly do all the majority classes. The clear distinction between minority classes and majority classes is particularly useful in revealing the performance improvement of the minority classes.

The long-tailed imbalance (Cui et al., 2019) is another type of imbalance, which follows the exponential decay. Specifically, the sample size in the $i$-th class, $N_i$, is chosen as $N_i = N_{max} \times \mu^i$, where $N_{max}$ is the sample size of the most frequent class (corresponding to $i = 0$) and $\mu < 1$ is the decay constant. We set $\mu = 0.85$ to obtain a highly imbalanced distribution in this work. In contrast with the step imbalance, the long-tailed imbalance has a smooth transition from majority to minority classes, which is presumably similar to the real-world data distribution.

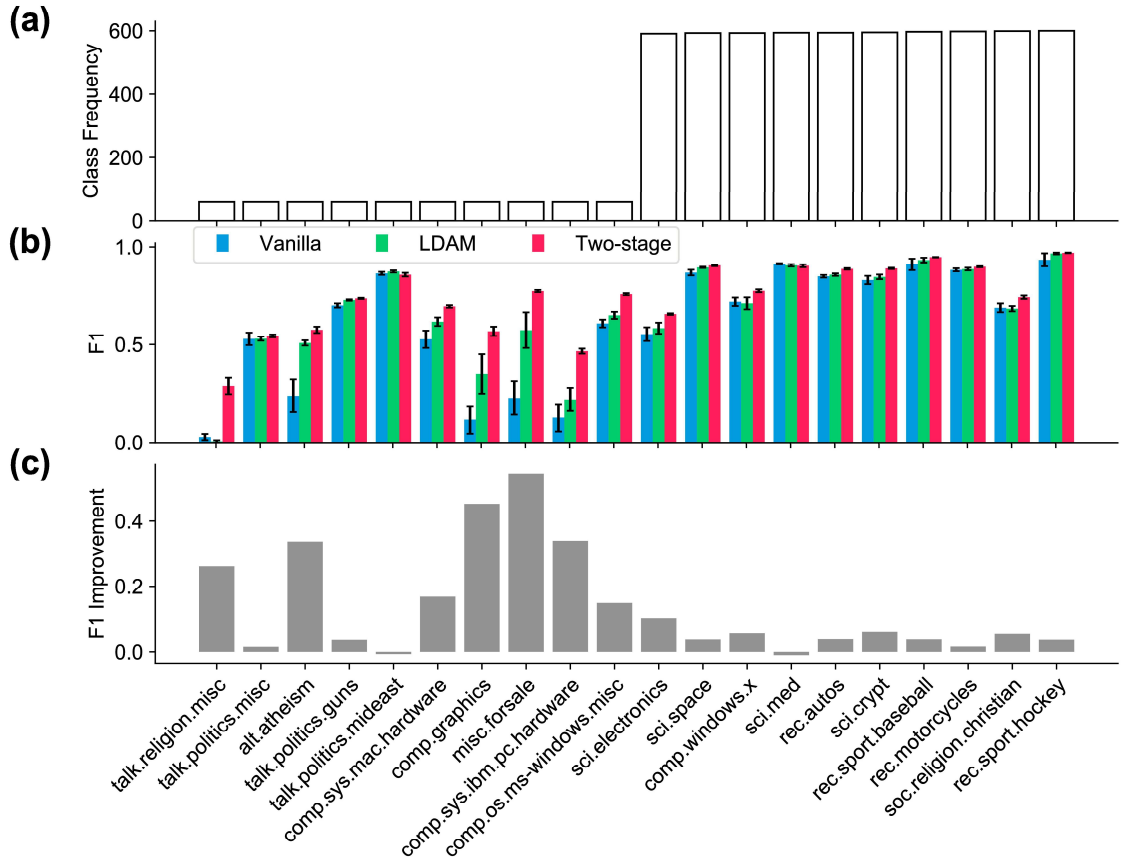

**Fig.4** Performance comparison between the two baseline methods (Vanilla and LDAM methods) and our two-stage fine-tuning on the artificially created step-imbalanced 20 Newsgroups dataset. **a** Step imbalance distribution of the training set. **b** Per-class F1-score comparison among three methods. **c** The F1 improvement of our two-stage fine-tuning over the vanilla method. In most of the minority classes, a large improvement is observed for our two-stage fine-tuning. Error bars denote the standard error.

### 4.3.2 Results

**Step imbalance**. The distribution of the synthetic step-imbalanced training data is presented in Figure 4(a). Figure 4(b) shows the performance for different methods for this step-imbalanced data, whereas the F1 improvement of our two-stage fine-tuning over the vanilla method is provided in Figure 4(c). The total micro F1 was 0.67674 for the vanilla fine-tuning, 0.715 for LDAM fine-tuning, and 0.76745 for the two-stage fine-tuning. Based on Figure 4, we observe that 9 out of 10 minority classes, our method performs better than both the vanilla and LDAM methods. The only minority class in which our method does not have the best F1-score is *talk.politics.mideast*, where all three methods perform almost equally well. Importantly, the minority classes gain much more improvement than the majority classes, indicating that our two-stage fine-tuning is indeed effective to cope with the imbalanced data.

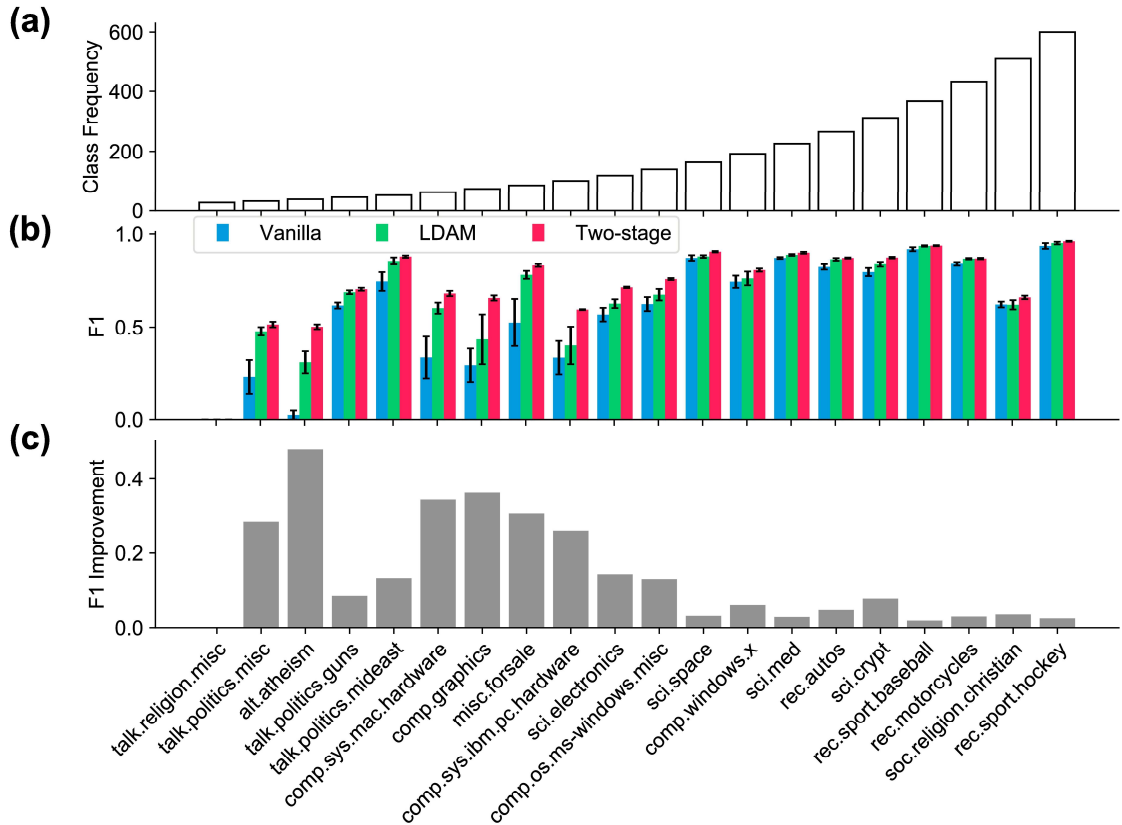

**Fig.5** Performance comparison between the two baseline methods (Vanilla and LDAM methods) and our two-stage fine-tuning on the artificially created long-tailed imbalanced 20 Newsgroups dataset. **a** Long-tailed imbalance distribution of the training set. **b** Per-class F1-score comparison among three methods. **c** The F1 improvement of our two-stage fine-tuning over the vanilla method. In most of the minority classes, a large improvement is observed for our two-stage fine-tuning. Importantly there are more improvements observed in the less frequent classes. Error bars denote the standard error.

**Long-tailed imbalance**. To check if our method can also be applied to a different imbalanced distribution, we use the long-tailed imbalance of training data, as shown in Figure 5(a). Results of training the model with this imbalanced data for three different methods, namely vanilla fine-tuning, LDAM finetuning, and two-stage fine-tuning are presented in Figure 5(b). In addition, Figure 5(c) shows the F1 improvement of two-stage fine-tuning over the vanilla method. We can see that our two-stage fine-tuning outperforms the other two methods in most classes. Similar to the step imbalance, more improvements are obtained among the minority classes. For the least frequent class of only 27 samples, talk.religion.misc, all three methods have zero F1 score. This is due to the rather small sample size, which is likely not sufficient for any of the classifiers to learn this class. Micro F1 was 0.6624 for the vanilla fine-tuning, 0.72114 for LDAM fine-tuning, and 0.7671 for the two-stage fine-tuning. As an independent verification, we additionally tested other values of μ to provide different exponential decays in the long-tailed imbalance, they all produce similar results (not shown for the sake of brevity).

## 4.4. Real-World Application: ADME Semantic Labeling

### 4.4.1 ADME Dataset

For this real-world application, we applied the proposed method to our curated FDA drug labeling dataset for enhancing product-specific guidances (PSGs)[2] assessment (Shi et al., 2022b). Recommended by the United States Food and Drug Administration (FDA), PSGs are instrumental to promote and guide generic drug product development. The FDA assessor needs to take extensive time and effort to manually retrieve supportive drug information of absorption, distribution, metabolism, and excretion (ADME) from the reference listed drug labeling for the PSG assessment. As a result, it is highly desirable to automate this process by developing a text classification model to automatically label ADME paragraphs with their semantic meaning. However, the pharmacokinetics section of the FDA-approved drug labeling contains not only the ADME paragraphs but also other topics, such as "specific populations", "drug interaction studies". By considering all topics other than ADME as "Other", the dataset is heavily class-imbalanced by nature.

We retrieved the FDA drug labeling from the DailyMed[3], which is a free drug information resource provided by the U.S. National Library of Medicine (NLM). The electronic drug labeling in DailyMed follows the Structured Product Labeling (SPL) standard, which specifies various drug label sections by Logical Observation Identifiers Names and Codes (LOINC). ADME is a part of the pharmacokinetics section (LOINC code: 43682-4) of drug labeling. We used the rule-based method to extract 5,687 ADME paragraphs with explicit ADME titles and 5,367 paragraphs under other topics (e.g., "specific populations", "drug interaction studies", etc.) and hence labeled them as "Other" from the pharmacokinetics section in drug labeling. For details about data preparation, please refer to (Shi et al., 2021). This dataset is imbalanced by nature since "Other" includes multiple topics in pharmacokinetics. We randomly split 85% of the dataset for training and the rest 15% for testing, so both training and testing datasets remained class-imbalanced.

---

[2] https://www.accessdata.fda.gov/scripts/cder/psg/index.cfm
[3] https://dailymed.nlm.nih.gov/dailymed/

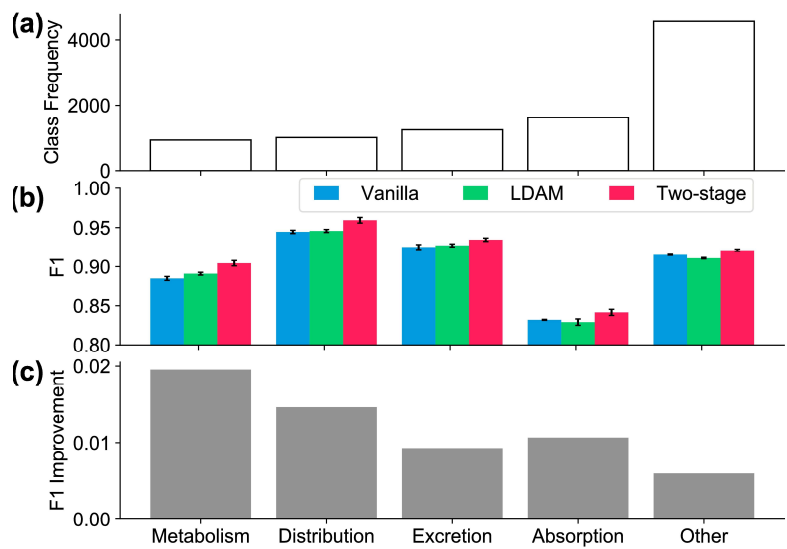

**Fig.6** Per-class F1-scores and improvements on ADME dataset. **a** The class distribution of the training set. **b** The per-class F1-score comparison between the two baseline methods (Vanilla and LDAM) and our two-stage fine-tuning. **c** Improvement in per-class F1 of our two-stage fine-tuning over the vanilla method. The F1-score increase is observed in all classes with our two-stage fine-tuning. The improvement becomes larger in the less frequent classes and smaller in the more frequent classes. Error bars denote the standard error.

### 4.4.2 Results

The distribution of class frequency in the training set is shown in Figure 6(a), where we can see the data is heavily class-imbalanced. Figure 6(b) shows the comparison of the F1-score among three methods. We observe that, for all classes, our two-stage fine-tuning has the best performance, and the LDAM performs similarly to the vanilla method. The per-class F1 improvement of our two-stage fine-tuning over the vanilla method is shown in Figure 6(c). A general trend can be observed from Figure 6(c): the F1 improvement decreases as the sample size (or class frequency) increases across all the classes except the Absorption class which has the lowest F1 score among all the classes. We note that Absorption is the passage of a drug travelling from the site of administration to the site of action (Yang et al., 2017), which therefore could be conflated with the Distribution that describes the reversible transfer of a drug from one location to another within the body. Micro F1 was 0.90211 for the vanilla fine-tuning, 0.90006 for LDAM fine-tuning, and 0.91163 for the two-stage fine-tuning. Overall, these results indicate that the simple modification to the standard fine-tuning in our proposed method indeed has a stronger effect on the less frequent classes to counteract the class imbalance in the data.

## 5. Discussion and Conclusion

Real-world classification problems typically exhibit a long-tailed or imbalanced distribution, wherein many classes are associated with only a few samples. This poses a challenge for learning on such data as it is susceptible to an undesirable bias towards majority classes, hence leading to poor performance for minority classes. To tackle this issue, we propose a new learning strategy: two-stage fine-tuning. In Stage 1, we fine-tune the model with a class-balanced reweighing loss to provide the minority classes a head-start. In Stage 2, we resume the standard fine-tuning. We compare our two-stage fine-tuning with standard fine-tuning and the LDAM fine-tuning and show that our two-stage fine-tuning outperforms both methods on two manually created imbalanced benchmark datasets (both two-class and multi-class) and a real-world application of ADME semantic labeling. Critically, we show that our two-stage fine-tuning possesses a better generalization ability when tested on OOD data.

Fine-tuning has become the *de facto* standard to leverage large pretrained models to perform downstream tasks (Devlin et al., 2019; Radford et al., 2019; Raffel et al., 2020). Despite widely used, fine-tuning may predispose the pretrained model to overfitting and generalizing poorly to OOD data due to the huge model and relatively small size of the domain-specific data. To address this gap, it has been shown (Shi et al., 2022a) that the pre-finetuning, as an intermediate training stage in between the pretrained model and fine-tuning, is effective to yield better performance. To put it in perspective, we may consider the first stage in our two-stage fine-tuning approach as the pre-finetuning. As such, our method entails all the benefits of the pre-finetuning. Having such an additional training stage, it encourages the pretrained model to be better adapted to the target data before the vanilla fine-tuning takes place.

In our two-stage fine-tuning approach, we have used the LDAM (Cao et al., 2019) for reweighting in our first stage due to its state-of-the-art performance in learning imbalanced data. We note, however, other reweighting schemes can also be used. The key for using reweighting in the first stage is to recompense the minority classes in order to protect the learning of minority classes from being at a disadvantage during the model updating and learn a good initial representation of the specific downstream task. In an independent and concurrent work, Kumar et al. (2022) show that a combined strategy of linear probing then full fine-tuning provides better results than each component alone. Our work, albeit conceptually similar, has at least three major differences: (1) we aim at the class imbalance problem; (2) we additionally fine-tune the final layer of the model, as motivated by theoretical analysis (Fang et al. 2021); and (3) we focus on the NLP tasks. We expect future work will continue exploring how our method compares to their method (Kumar et al., 2022) and different reweighting schemes including deferred re-weighting (Cao et al., 2019).

When reporting the model performance, it has been an on-going debate about how to choose an appropriate metric for imbalanced data (Japkowicz, and Shah, 2011), particularly for multi-class classification problems. In this work, we have mainly used the F1-score to assess the model performance, though other measures can be more appropriate in different domains such as medical diagnosis. As the F-1 score is more sensitive to data distribution, it is a suitable measure for classification problems on imbalanced datasets. As we work with both two-class and multi-class problems, we report both the overall F-1 score (micro-F1) and per-class F-1 score to quantify the generalization performance of both majority and minority classes for the benchmark datasets. We also test the per-class top-1 error, which obtains similarly consistent performance. When reporting the model performance for imbalanced data, we deem that it is important to explicitly document what evaluation metrics are exactly used.

There are several extensions of our method. First, the effectiveness of our two-stage fine-tuning is primarily demonstrated on text classification tasks in this work. It is conceivable that our method can be equally applied to other domains of machine learning such as computer vision, where the imbalanced data abound. Second, while the method we proposed in this work is designed for learning imbalanced data, the idea can also be beneficial for balanced data. Third, in the current work, we have limited the fine-tuning at the first stage to only the final layer, which is mainly determined by the theoretical analysis (Fang et al., 2021). In practice, it is plausible that the last few layers, rather than only the final layer, actually can be fine-tuned for improved performance when it comes to different datasets in real-world applications. Finally, our work focuses on the imbalanced classification. Extending our investigation to imbalanced regression (Torgo et al., 2013; Steininger et al., 2021; Yang et al., 2021) is also of interest.

**Acknowledgments**: Work reported here was run on hardware supported by Drexel's University Research Computing Facility.

**Author Contributions**: Conceptualization: T.V. and H.L.; Methodology: T.V. and H.L; Formal analysis and investigation: T.V., Y.S., and H.L.; Writing – Original Draft, T.V. and H.L.; Writing – Review & Editing, T.V., Y.S., M.H., L.Z., and H.L..; Funding Acquisition: H.L.; Resources: J.W., P.R., Y.Z., M.H, and L.Z..; Supervision: H.L.

**Funding**: This work was partly supported by The United States Food and Drug Administration Contract #: 75F40119C10106.

**Conflict of interest**: The authors declare that they have no conflict of interest.

**Disclaimer**: The opinions expressed in this article are the author's own and do not reflect the view of the Food and Drug Administration, the Department of Health and Human Services, or the United States government.